\documentclass[final]{cvpr}
\usepackage{times}
\usepackage{epsfig}
\usepackage{graphicx}
\usepackage{amsmath}
\usepackage{amssymb}
\usepackage{dsfont}
\usepackage{array}
\usepackage{booktabs}
\usepackage[pagebackref=true,breaklinks=true,colorlinks,bookmarks=false]{hyperref}
\def\cvprPaperID{2532} 
\def\confYear{CVPR 2021}

\begin{document}

\title{Non-Salient Region Object Mining for Weakly Supervised Semantic Segmentation}

\author{Yazhou Yao$^{1}$, 
	    Tao Chen$^{1}$,
	    Guo-Sen Xie$^{2}$, 
	    Chuanyi Zhang$^{1}$,\\ 
	    Fumin Shen$^{3}$, 
	    Qi Wu$^{4}$,  
	    Zhenmin Tang$^{1}$, 
	    Jian Zhang$^{5}$  \\
$^{1}$Nanjing University of Science and Technology; $^{2}$Mohamed bin Zayed University of AI\\
$^{3}$University of Electronic Science and Technology of China; $^{4}$University of Adelaide;\\ $^{5}$University of Technology Sydney\\
}

\maketitle

\pagestyle{empty}  
\thispagestyle{empty} 

\begin{abstract}
Semantic segmentation aims to classify every pixel of an input image.  Considering the difficulty of acquiring dense labels, researchers have recently been resorting to weak labels to alleviate the annotation burden of segmentation. However, existing works mainly concentrate on expanding the seed of pseudo labels within the image's salient region. In this work, we propose a non-salient region object mining approach for weakly supervised semantic segmentation. We introduce a graph-based global reasoning unit to strengthen the classification network's ability to capture global relations among disjoint and distant regions. This helps the network activate the object features outside the salient area. To further mine the non-salient region objects, we propose to exert the segmentation network's self-correction ability. Specifically, a potential object mining module is proposed to reduce the false-negative rate in pseudo labels. Moreover, we propose a non-salient region masking module for complex images to generate masked pseudo labels. Our non-salient region masking module helps further discover the objects in the non-salient region. Extensive experiments on the PASCAL VOC dataset demonstrate state-of-the-art results compared to current methods. The source codes are available at \url{https://github.com/NUST-Machine-Intelligence-Laboratory/nsrom}.
\end{abstract}

\section{Introduction}

Semantic segmentation is the task of classifying every pixel of an input image. It plays a vital role in many computer vision tasks, such as image editing and medical image analysis \cite{2020Group,wang2021exploring}. Benefiting from the recent advances of deep learning, semantic segmentation has achieved remarkable progress. However, the training of deep convolutional neural networks (CNNs) usually requires large-scale datasets \cite{yao2018extracting,yao2020exploiting,yao2019towards,yao2018extractingtip,yao2017exploiting}. Moreover, obtaining precise pixel-wise annotations  for semantic segmentation demands intensive labor efforts and is quite time-consuming. One promising approach to address the annotation problem for semantic segmentation is to learn from weak labels, such as image-level annotations \cite{kolesnikov2016seed,wei2016stc,hong2017weakly,chaudhry2017discovering,huang2018weakly,ahn2018learning,wei2018revisiting,jiang2019integral}, bounding boxes \cite{dai2015boxsup,khoreva2017simple,song2019box}, points \cite{bearman2016s}, and scribbles \cite{lin2016scribblesup,vernaza2017learning}. Among these weak supervisions, image-level labels are the easiest format to annotate and have been widely studied in various weakly supervised methods. However, semantic segmentation supervised with image-level labels remains a challenging task. Therefore, this paper follows the current trend and focuses on leveraging image-level labels to achieve weakly supervised semantic segmentation (WSSS).

\begin{figure}[t]
\begin{center}
\includegraphics[width=\linewidth]{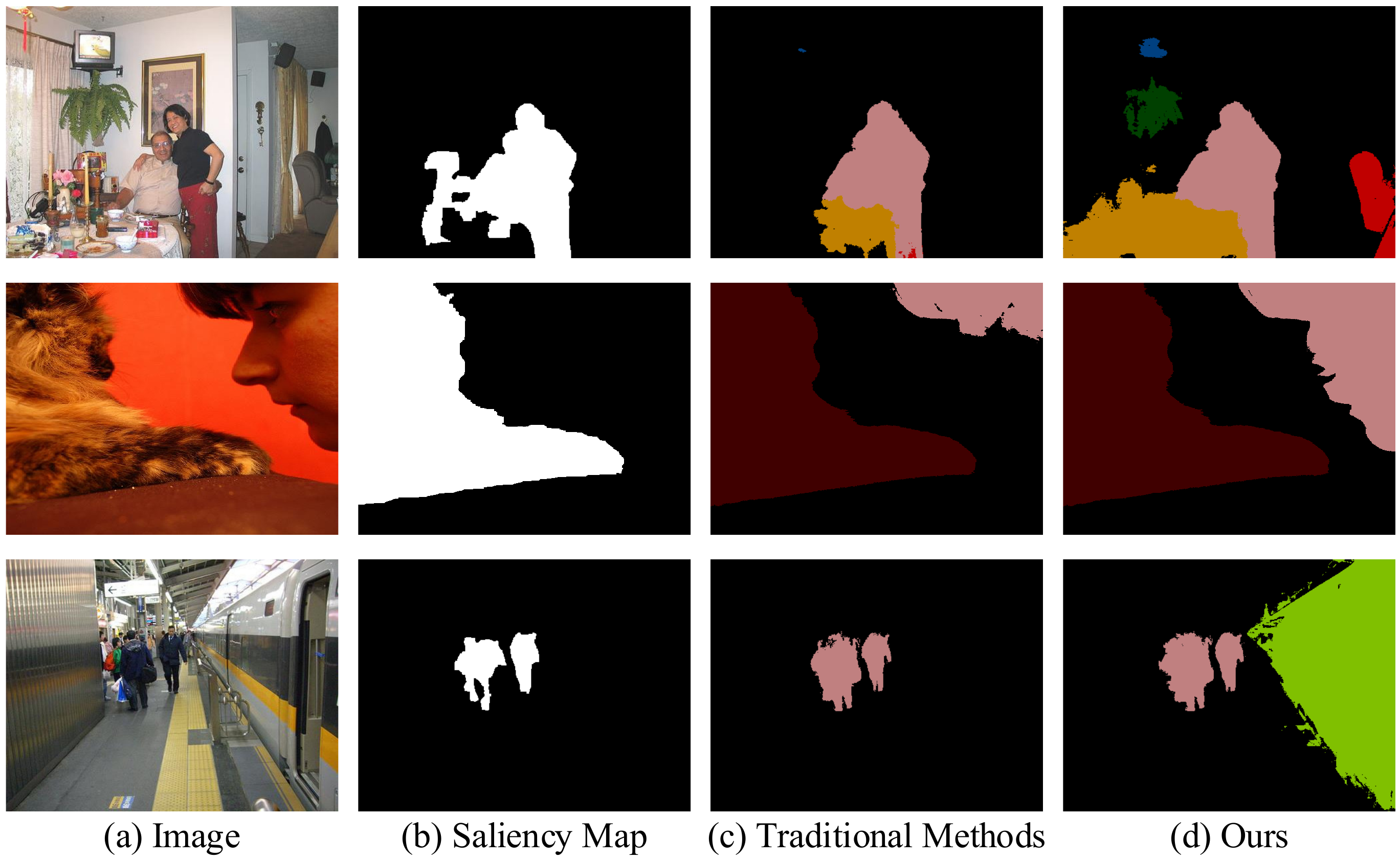}
\end{center}
\vspace{-0.15cm}
\caption{Comparison between the traditional methods and ours. (a) Input image. (b) The saliency map. (c) Results of the traditional methods. They mainly focus on expanding the seed of pseudo labels within the salient region of the image. They only obtain good segmentation results in conspicuous regions. (d) Our results. Our method also mines the object in the non-salient region and can get better results both in and outside the salient region. Best viewed in color.}
\vspace{-0.2cm}
\label{fig_moti}
\end{figure}

To tackle the task of WSSS with only image-level labels, visualization-based approaches \cite{zhou2016learning} have been widely adopted to narrow the annotation gap between classification and segmentation \cite{yao2020bridging,yao2016domain}. The typical methods train a classification network with image-level labels. Then they leverage class activation maps (CAMs) \cite{zhou2016learning} to generate pseudo labels to train the segmentation network. However, these activation maps obtained from the classification network are sparse and incomplete. They can only locate the most discriminative part of objects. Many approaches have been proposed to enlarge the activated region to cover a large object area. For example, Jiang \etal \cite{jiang2019integral} observed that the attention maps produced by the classification network focus on different object parts during training. Therefore, they proposed an online attention accumulation (OAA) strategy to combine the various activated regions. However, as shown in Fig.~\ref{fig_moti}, the existing works mainly concentrate on enlarging the response maps for the salient region. Then they utilize the saliency maps to extract background. Few works focus on mining objects in the non-salient areas.

In this paper, we propose a non-salient region object mining method for weakly supervised semantic segmentation to make up for the shortcomings mentioned above. In contrast to the widely adopted center prior \cite{borji2015salient} for saliency detection, the non-salient region is usually scattered in corners or near the edge of the image. Such a characteristic of our protagonist requires the network to exploit the disjoint and distant surrounding information. While the traditional classification networks based on CNNs excel at modeling local relations, they are inefficient at capturing global relations between disjoint and distant regions. Therefore, we introduce a graph-based global reasoning unit \cite{chen2019graph} to strengthen the classification network's capability in activating the object features outside the salient region. 

On the other hand, though existing approaches can successfully enlarge activated regions for objects, they inevitably extend the object area to the background. These methods require the saliency maps to provide background clues. While the saliency maps can correct the pixel labels near conspicuous regions, they also remove the object labels outside the salient area. We notice that although the naive CAM, sparse and incomplete, does not have an accurate boundary, it can provide useful clues for the objects in the non-salient region. Therefore, we propose a potential object mining module to discover more objects that are outside the conspicuous region but activated in the naive CAM. Our potential object mining module aims to reduce the pseudo labels' false-negative rate (in which case the object regions are falsely labeled as background). This improves the quality of pseudo labels and encourages the segmentation network to exert its self-correction ability. Such an ability of the network inspires us to further take advantage of the prediction of the segmentation network. Following \cite{wei2016stc}, we divide the training images into simple and complex sets according to the number of categories in each image. The simple images with a single category of object(s) usually have a clean background. Their objects often exist in the conspicuous region and can be correctly segmented. In contrast, complex images (having two or more categories of objects) are more prone to having objects outside the salient area. Therefore,  we propose a non-salient region masking module for complex images to generate masked pseudo labels. Our non-salient region masking module helps further discover objects in the non-salient region. Our contributions can be summarized as follows:

\begin{itemize}
	
	\item For weakly supervised semantic segmentation, we leverage a global reasoning unit to capture global relations among disjoint and distant regions, helping the network activate object features outside salient areas.
	
	\item We propose a potential object mining module to discover more objects in the non-salient region, which improves the quality of pseudo labels by reducing the false-negative rate. 
	
	\item We propose a non-salient region masking module with a dilation policy to generate masked pseudo labels, which leads to a more robust segmentation model to further discover objects outside the salient region.
	
\end{itemize}

\section{Related Work}

\subsection{Semantic Segmentation}
Semantic segmentation is an important computer vision task that assigns a semantic label to every pixel in an image. Since the adaption of the modern classification network into the fully convolutional network (FCN) \cite{long2015fully,sun2020crssc}, deep learning has achieved great success in semantic segmentation \cite{badrinarayanan2017segnet,chen2017deeplab,zhang2018context,liu2019auto,chen2020classification,chen2021semantically,luo2019segeqa}. To address the size issue caused by the down-sampling operation, early works \cite{badrinarayanan2017segnet} adopted an encoder-decoder architecture to recover the spatial resolution. Then dilated/atrous convolution \cite{chen2017deeplab} was proposed for the expansion of the receptive field without loss of resolution. Recently, the pyramid pooling module and context encoding \cite{zhang2018context} were introduced to capture the global semantic context of the scene. Auto-DeepLab \cite{liu2019auto} presented a network-level search space to allow efficient gradient-based architecture search for semantic segmentation.

\begin{figure*}[t]
\begin{center}
\includegraphics[width=\linewidth]{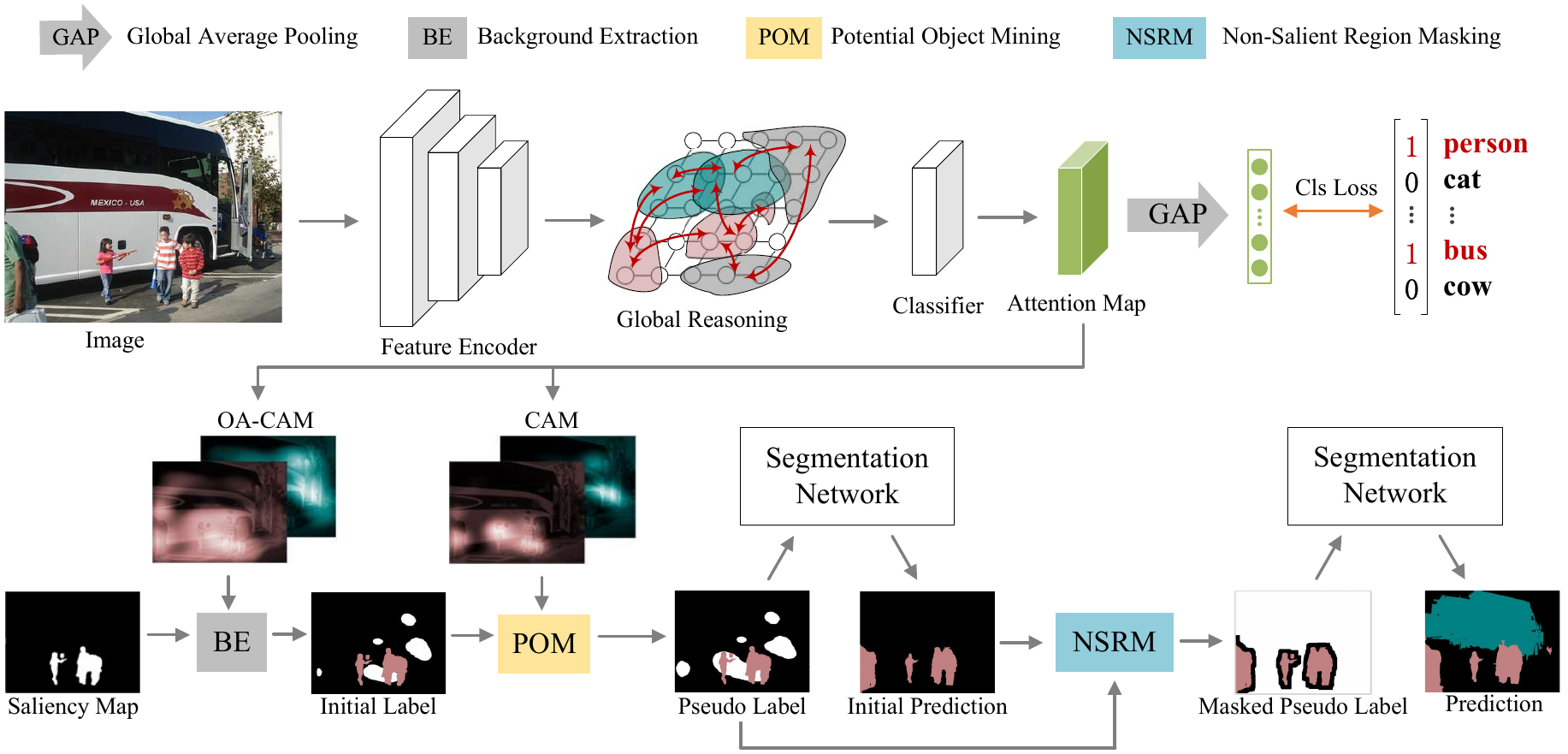}
\end{center}
\vspace{-0.15cm}
	\caption{The architecture of our proposed approach. We train a classification network to generate the class activation maps (CAMs) and online accumulated class attention maps (OA-CAMs). A graph-based global reasoning unit is inserted into the classification network to help activate the objects outside the salient region. After obtaining the initial label with background extraction (BE), we utilize the potential object mining module (POM) to discover more objects missed in the initial label. We further leverage the non-salient region masking module (NSRM) to generate masked pseudo labels for the training of the segmentation network. Best viewed in color.}
	\label{fig_framework}
	\vspace{-0.2cm}
\end{figure*}

\subsection{Weakly Supervised Semantic Segmentation}

Weakly supervised semantic segmentation attempts to learn a segmentation network with weaker annotation than pixel-wise labels. It aims to alleviate the annotation burden of segmentation tasks. Compared to bounding boxes \cite{dai2015boxsup,khoreva2017simple,song2019box}, points \cite{bearman2016s}, and scribbles \cite{lin2016scribblesup,vernaza2017learning}, image-level labels \cite{kolesnikov2016seed,wei2016stc,hong2017weakly,chaudhry2017discovering,huang2018weakly,ahn2018learning,wei2018revisiting,jiang2019integral} are the most widely used weak annotations due to their easy availability. They are already given in existing large-scale datasets (\eg ImageNet \cite{deng2009imagenet}) or can be automatically generated through image retrieval techniques. Existing image-level label based approaches leverage the CAM to generate pixel-level seeds for training the segmentation model. Considering the initial seeds' sparsity and incompleteness, researchers proposed many approaches to expand the seeds to integral object regions. For example, Kolesnikov and Lampert \cite{kolesnikov2016seed} introduced a new loss function for weakly supervised training based on three guiding principles: seed, expand, and constrain. Huang \etal \cite{huang2018weakly} proposed to train a model starting from the discriminative regions and progressively increase the pixel-level supervision with the deep seeded region growing strategy. The work of RDC \cite{wei2018revisiting} leveraged the dilated convolution to enlarge the receptive fields of convolutional kernels. This helped transfer the object information to the non-discriminative region. AffinityNet \cite{ahn2018learning} realized semantic propagation by a random walk with the semantic affinity between a pair of adjacent image coordinates. The recent work of SEAM \cite{wang2020self} proposed a self-supervised equivariant attention mechanism to provide additional supervision for network learning. Apart from using intra-image information, Sun \etal \cite{sun2020mining} incorporated two neural co-attentions into the classifier to capture cross-image semantic relations for comprehensive object pattern mining. Zhang \etal \cite{zhang2020causal} attributed the reason for the ambiguous boundaries of pseudo-masks to the confounding context. They presented a causal inference framework to remove the confounding bias in image-level classification with an effective approximation for the backdoor adjustment.

\section{The Proposed Approach}
In this paper, we focus on the task of weakly supervised semantic segmentation with image-level labels. Our framework is illustrated in Fig.~\ref{fig_framework}. Given a set of training images with image-level labels, we train a classification network. We leverage class activation maps to generate pseudo labels for learning a segmentation network. Unlike existing methods that mainly concentrate on refining pseudo labels in the salient area, we propose to discover more objects in the non-salient region for weakly supervised semantic segmentation. To achieve this, we insert a graph-based global reasoning unit into the classification network. This helps to activate the object features outside the salient region. We also adopt a potential object mining module (POM) and a non-salient region masking module (NSRM) to improve the quality of pseudo labels for non-salient region object mining.

\subsection{CAM Generation}

A classification network is first trained to generate class attention maps. As illustrated in Fig.~\ref{fig_framework}, to strengthen the classification network's ability to capture global relations among disjoint and distant regions, we introduce a graph-based global reasoning unit \cite{chen2019graph} before the final classifier. The global reasoning module will help the network to activate the object parts outside the salient region. The features $X\in \mathds{R}^{L\times K}$ generated by the encoder, with $K$ being the feature dimension and $L= H\times W$ locations, is first projected from the coordinate space to a latent interaction space. The projection function $V = f\left ( X \right )\in \mathds{R}^{N\times K}$ is formulated as a linear combination:
\begin{equation}
	v_{i}=b_{i}X = \sum_{j}b_{ij}x_{j},
\end{equation}
where $B = \left [ b_{1},\cdots ,b_{N} \right ]\in \mathds{R}^{N\times L}$ is the learnable projection weights, and $N$ is the number of the features (nodes) in the interaction space.

Then a graph convolution \cite{kipf2016semi} is applied to capture the relations between features in the new space:
\begin{equation}
	Z = \left (\left ( I-A_{g} \right )V  \right )W_{g}.
\end{equation}
$A_{g}$ denotes the $N\times N$ node adjacency matrix learned by gradient descent during training. $W_{g}$ denotes the state update function.

After obtaining the node-feature $Z\in \mathds{R}^{N\times K}$, reverse projection $Y = g\left ( Z \right )\in \mathds{R}^{L\times K}$ is conducted to project the feature back to the original space:
\begin{equation}
	y_{i}=d_{i}Z = \sum_{j}d_{ij}z_{j},
\end{equation}
where $D = \left [ d_{1},\cdots ,d_{N} \right ] = B^{T}$.

For the training of the classification network, we adopt the multi-label soft margin loss as follows:
\begin{equation}
	L_{cls}=-\frac{1}{C} \sum_{c=1}^{C} y_{c} \log \sigma\left(p_{c}\right)+\left(1-y_{c}\right) \log \left[1-\sigma\left(p_{c}\right)\right].
\end{equation}
Here, $p_{c}$ is the prediction of the network for the $c$-th class. $\sigma\left(\cdot \right)$ is the sigmoid function, and $C$ is the total number of foreground classes. $y_{c}$ is the image-level label for the $c$-th class. Its value is 1 if the class is present in the image; otherwise, its value is 0.

We obtain CAMs by selecting the class-specific feature maps generated by the final classifier. Following OAA \cite{jiang2019integral}, we also generate the online accumulated class attention maps (OA-CAMs) to have more entire regions and strengthen the lower attention values of target object regions with their integral attention model.

\subsection{Potential Object Mining}
After obtaining OA-CAMs, the work of OAA \cite{jiang2019integral} uses them to extract object cues and saliency maps to extract background cues. The class label of each pixel is assigned by comparing the value of each OA-CAM. As shown in Fig.~\ref{fig_framework}, with the shape information provided by the saliency map, the initial label is derived with quite clear object boundaries after the background extraction (BE) process. However, the initial label misses many object parts outside the conspicuous area. Therefore, we propose to discover more objects in the non-salient region. Though the OA-CAM has a high recall of the object pixel, its precision is low. In contrast, the CAM, widely leveraged to generate initial seeds for proxy segmentation labels \cite{kolesnikov2016seed,huang2018weakly}, has low recall but high precision. Therefore, we propose a potential object mining (POM) module to discover the object region activated in the CAM. We mine the potential object with a class adaptive threshold $T_{c}$ for class $c$ that is present in the image:

\begin{equation}
T_{c} = \left\{
\begin{array}{ll}
MED(v)  & \text {if}\  \exists \left (i,j  \right ),\mathrm{s.t.}\  l_{ij}=c \\
TQ(v)  & \text { otherwise }
\end{array}.
\right.
\label{eq_t}
\end{equation}
Here, v is the set of attention values of pixels in the CAM, whose locations $p$ are selected as follows:
\begin{equation}
p = \left\{
\begin{array}{ll}
 \left \{ (i,j)|l_{ij}=c \right \} & \text {if}\ \exists \left (i,j  \right ),\mathrm{s.t.}\l_{ij}=c \\
 \left \{ (i,j)|a_{ij}>T_{bg} \right \} & \text { otherwise }
\end{array},
\right.
\label{eq_p}
\end{equation}
where $a_{ij}$ is the attention value in CAM at the position (i,j). $l_{ij}$ is the value in the initial label at the position (i,j), which denotes the pseudo label of the pixel. As illustrated in Equation~\ref{eq_t} and Equation~\ref{eq_p}, if the initial label contains class $c$, we select those pixels in its CAM and choose the median (MED) of their attention values as $T_{c}$. Otherwise, we select pixels in its CAM with an attention value greater than the background threshold $T_{bg}$ and choose the top quartile (TQ) of their attention values as $T_{c}$. 

We then adjust the initial label as follows:
\begin{equation}
l_{ij} = \left\{
\begin{array}{ll}
255 & \text {if}\ \exists \ c, \left (i,j  \right ), \mathrm{s.t.}\ l_{ij} = 0, a_{i,j}^{c}>T_{c} \\
l_{ij} & \text { otherwise }
\end{array}.
\right.
\label{eq_l}
\end{equation}
Here, $a^{c}$ denotes the CAM for class $c$. As illustrated in Equation~\ref{eq_l}, the background pixels (labeled as 0) in the initial label with any CAM attention value greater than $T$ are labeled as 255 and ignored for training. We do not label them for the corresponding potential class to avoid introducing wrong object labels. Such a strategy bypasses the necessity to locate the object boundary outside the salient region. We focus on reducing the false-negative rate of pseudo labels, which will help discard the gradients generated by the misleading information.

\subsection{Non-Salient Region Masking}
Our potential object mining strategy enriches pseudo labels with more ignored pixels. It allows the segmentation network to predict the correct labels for these potential object regions during training. The improved quality of pseudo labels can also encourage the segmentation network to fix the other incorrectly labeled regions. Therefore, we propose to further leverage the prediction of the segmentation model to generate pseudo labels of higher quality for retraining.    

We notice that simple images with only one category of objects usually have a clean background. Objects in these images often exist in the salient region and can be correctly segmented by the segmentation network. However, complex images (with two or more categories of objects) are more prone to having objects outside the salient area. It remains challenging for the segmentation network to detect objects outside the salient region with pseudo labels only containing object labels in the salient area. Therefore, we propose a non-salient region masking (NSRM) module. It combines the object information in the segmentation network's prediction and pseudo labels to generate masked labels for complex images.

\begin{figure}[t]
	\begin{center}
		\includegraphics[width=\linewidth]{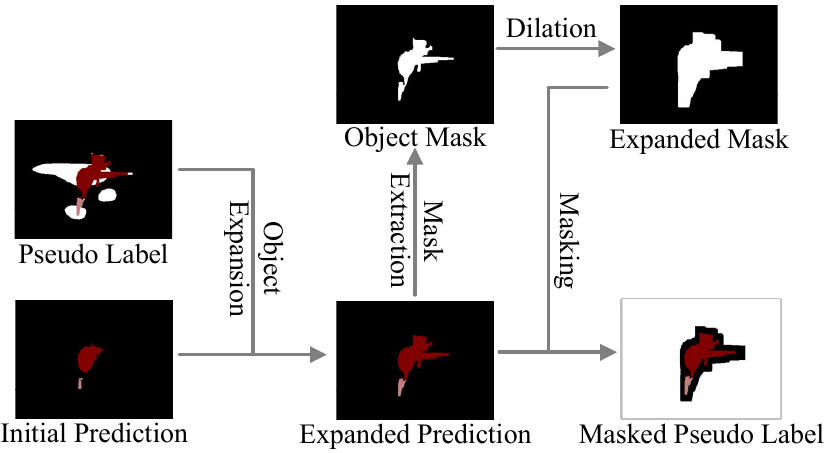}
	\end{center}
\vspace{-0.15cm}
	\caption{Our proposed non-salient region masking module.}
	\vspace{-0.2cm}
	\label{fig_nsrm}
\end{figure}

Our proposed non-salient region masking module is illustrated in Fig.~\ref{fig_nsrm}. Based on the assumption that object labels within the salient region are correct with high probability, we first expand the object region in the initial prediction with the guidance of our pseudo labels. Then we extract the object mask from the expanded prediction map. After that, we expand the object mask with a dilation operation. Finally, a masking operation is applied to the expanded prediction map to get the masked pseudo labels. Note that the dilation operation introduces a small portion of the background around the objects. It preserves the objects' boundary information, which is of great importance for a successful segmentation network.

\section{Experiments}
\subsection{Implementation Details}
For the classification network, we adopt the VGG-16 model as our backbone, which is pre-trained on ImageNet \cite{deng2009imagenet}. Following \cite{jiang2019integral}, we add three convolutional layers on the top of the fully-convolutional backbone. A ReLU layer follows each convolutional layer for nonlinear transformation. A $1\times1$ convolutional layer of $C$ channels is adopted as the pixel-wise classifier to generate attention maps. The momentum and weight decay of the SGD \cite{bottou2010large} optimizer are 0.9 and $5 \times 10^{-4}$. The initial learning rate is set to $10^{-3}$ and is divided by 10 after every 5 epochs. Following the code of \cite{jiang2019integral}, we set the background threshold $T_{bg}$ = 0.3 for fair comparison. We train the classification network for 14 epochs with batch size = 5.

For the segmentation network, following \cite{chang2020weakly,zhang2020splitting,fan2020employing,chen2020weakly}, we adopt the DeepLab-v2 \cite{chen2017deeplab} framework. VGG-16 is pre-trained on ImageNet \cite{deng2009imagenet}. For ResNet-101 \cite{he2016deep}, we report results for models pre-trained on ImageNet \cite{deng2009imagenet} and MS-COCO \cite{lin2014microsoft}, respectively. The momentum and weight decay of SGD optimizer are 0.9 and $10^{-4}$. The initial learning rate is set to $ 2.5 \times 10^{-4}$ and is decreased using polynomial decay with a power of 0.9. The segmentation network is trained for 10,000 iterations with batch size = 10.

\subsection{Datasets and Evaluation Metrics}
Following previous works, we evaluate our approach on the Pascal VOC 2012 dataset \cite{everingham2010pascal}. It contains 21 classes (20 object categories and the background) for semantic segmentation. There are 10,582 training images, which are expanded by \cite{hariharan2011semantic}, 1,449 validation images, and 1,456 test images. For all the experiments, we only adopt the image-level class labels for training. Standard mean intersection over union (mIoU) is taken as the evaluation metric for the semantic segmentation task.

\subsection{Comparisons to the State-of-the-arts}
\textbf{Baselines.} In this part, we compare our proposed method with the following state-of-the-arts approaches that leverage image-level labels for weakly supervised semantic segmentation: DCSM \cite{shimoda2016distinct}, SEC \cite{kolesnikov2016seed}, AugFeed \cite{qi2016augmented}, STC \cite{wei2016stc}, Roy \etal \cite{roy2017combining}, Oh \etal \cite{oh2017exploiting}, AE-PSL \cite{wei2017object}, WebS-i2 \cite{jin2017webly}, Hong \etal \cite{hong2017weakly}, DCSP \cite{chaudhry2017discovering}, TPL \cite{kim2017two}, GAIN \cite{li2018tell}, DSRG \cite{huang2018weakly}, MCOF \cite{wang2018weakly}, AffinityNet \cite{ahn2018learning}, RDC \cite{wei2018revisiting}, SeeNet \cite{hou2018self}, OAA \cite{jiang2019integral}, ICD \cite{fan2020learning}, BES \cite{chen2020weakly}, Fan \etal \cite{fan2020employing}, Zhang \etal \cite{zhang2020splitting}, MCIS \cite{sun2020mining}, IRN \cite{ahn2019weakly}, FickleNet\cite{lee2019ficklenet}, SSDD \cite{shimoda2019self}, SEAM \cite{wang2020self}, SCE \cite{chang2020weakly}, CONTA \cite{zhang2020causal}.

\textbf{Experimental Results.} We present our results for the backbone of VGG and ResNet in Table \ref{tab_vgg} and Table \ref{tab_resnet}, respectively. As can be seen, our approach achieves better results than other state-of-the-art methods for both VGG and ResNet backbones. Specifically, for the VGG backbone, our segmentation results reach 65.5\% and 65.3\% on the validation and test set, respectively. For the ResNet backbone, we can get 68.3\% on the validation set and 68.5\% on the test set. Though the methods of STC \cite{wei2016stc},  WebS-i2 \cite{jin2017webly} and Hong \etal \cite{hong2017weakly} leverage additional training data, our method  outperforms them on the validation set by 15.7\%, 12.1\% and 7.4\%, respectively. Compared to DSRG \cite{huang2018weakly} and CONTA \cite{zhang2020causal}, which also utilize the prediction of the segmentation network to get refined pseudo labels for training, our approach can improve their results by 6.9\% and 2.2\%, respectively. The work of ICD \cite{fan2020learning} uses the additional superpixel to help recover the object boundary information during training. Our approach can still outperform it by 1.5\% for the VGG backbone and 0.5\% for the ResNet backbone. Our results demonstrate the effectiveness of mining the objects in the non-salient region for the task of weakly supervised semantic segmentation. When training the ResNet based network with the COCO pre-trained weights, we can further reach 70.4\% and 70.2\% on the validation and test set, respectively.

\begin{table}[t]
	
	\setlength{\tabcolsep}{2.8mm}
	\renewcommand\arraystretch{1.0}
	\centering
	\begin{tabular}{{l}*{3}{c}}
		\toprule
		Methods & Publication & Val & Test\\
		\midrule
		DCSM \cite{shimoda2016distinct} &ECCV16&44.1 &45.1\\ 
		SEC \cite{kolesnikov2016seed} &ECCV16&50.7 &51.7\\ 
		AugFeed \cite{qi2016augmented} &ECCV16&54.3 &55.5\\ 
		STC \cite{wei2016stc}&TPAMI17&49.8 &51.2\\ 
		Roy \etal \cite{roy2017combining} &CVPR17&52.8 &53.7\\ 
		Oh \etal \cite{oh2017exploiting} &CVPR17&55.7 &56.7\\ 
		AE-PSL \cite{wei2017object} &CVPR17&55.0 &55.7\\ 
		WebS-i2 \cite{jin2017webly}&CVPR17&53.4 &55.3\\ 
		Hong \etal \cite{hong2017weakly} &CVPR17&58.1 &58.7\\ 
		DCSP \cite{chaudhry2017discovering} &BMVC17&58.6 &59.2\\   
		TPL \cite{kim2017two} &ICCV17&53.1 &53.8\\ 
		GAIN \cite{li2018tell} &CVPR18&55.3 &56.8\\ 
		DSRG \cite{huang2018weakly} &CVPR18&59.0 &60.4\\
		MCOF \cite{wang2018weakly} &CVPR18&56.2 &57.6\\
		AffinityNet \cite{ahn2018learning} &CVPR18&58.4 &60.5\\  
		RDC \cite{wei2018revisiting} &CVPR18&60.4 &60.8\\
		SeeNet \cite{hou2018self} &NIPS18&63.1 &62.8\\
		OAA \cite{jiang2019integral}&ICCV19&63.1 &62.8\\  
		ICD \cite{fan2020learning} &CVPR20&64.0 &63.9\\
		BES \cite{chen2020weakly}&ECCV20&60.1 &61.1\\ 
		Fan \etal \cite{fan2020employing} &ECCV20&64.6 &64.2\\
		Zhang \etal \cite{zhang2020splitting} &ECCV20&63.7 &64.5\\  
		MCIS \cite{sun2020mining}&ECCV20&63.5 &63.6\\ 
		\textbf{Ours} &-&\textbf{65.5} &\textbf{65.3}\\	
		\bottomrule			
	\end{tabular}
\caption{\small{Quantitative comparisons to previous state-of-the-art approaches with VGG backbone.}}
\label{tab_vgg}	
\end{table}	

\begin{table}[t]
	
	\setlength{\tabcolsep}{2.8mm}
	\renewcommand\arraystretch{1.0}
	\centering
	\begin{tabular}{{l}*{3}{c}}
		\toprule
		Methods & Publication & Val & Test\\
		\midrule
		DCSP \cite{chaudhry2017discovering} &BMVC17&60.8 &61.9\\
		DSRG \cite{huang2018weakly} &CVPR18&61.4 &63.2\\ 
		MCOF \cite{wang2018weakly} &CVPR18&60.3 &61.2\\ 
		AffinityNet \cite{ahn2018learning} &CVPR18&61.7 &63.7\\
		SeeNet \cite{hou2018self} &NIPS18&63.1 &62.8\\
		IRN \cite{ahn2019weakly} &CVPR19&63.5 &64.8\\ 
		FickleNet\cite{lee2019ficklenet} &CVPR19&64.9 &65.3\\
		OAA \cite{jiang2019integral}&ICCV19&65.2 &66.4\\
		SSDD \cite{shimoda2019self}&ICCV19&64.9 &65.5\\   
		SEAM \cite{wang2020self} &CVPR20&64.5 &65.7\\
		SCE \cite{chang2020weakly} &CVPR20&66.1 &65.9\\
		ICD \cite{fan2020learning} &CVPR20&67.8 &68.0\\
		Zhang \etal \cite{zhang2020splitting} &ECCV20&66.6 &66.7\\
		Fan \etal \cite{fan2020employing} &ECCV20&67.2 &66.7\\
		MCIS \cite{sun2020mining} &ECCV20&66.2 &66.9\\
		BES \cite{chen2020weakly}&ECCV20&65.7 &66.6\\  
		CONTA \cite{zhang2020causal}&NIPS20&66.1 &66.7\\
		
		\textbf{Ours} &-&\textbf{68.3} &\textbf{68.5}\\	
		\midrule
		OAA* \cite{jiang2019integral}&ICCV19&67.4 &-\\
		SEAM* \cite{wang2020self} &CVPR20&63.2 &-\\
		\textbf{Ours}*  &-&\textbf{70.4} &\textbf{70.2}\\		
		\bottomrule	
	\end{tabular}
\caption{\small{Quantitative comparisons to previous state-of-the-art approaches with ResNet backbone. * denotes model is pre-trained on MS-COCO. }}
	\label{tab_resnet}	
\end{table}

\begin{table}[t]

	\setlength{\tabcolsep}{4mm}
	\renewcommand\arraystretch{1.0}
	\centering
	\begin{tabular}{{l}*{1}{c}}
		\toprule
		Methods  & Val \\
		\midrule
		baseline &67.7\\
		+ GR &68.8\\
		+ GR + retrain &68.7\\
		+ GR + POM &69.0\\
		+ GR + POM + retrain &69.7\\
		+ GR + POM + NSRM &70.4\\		
		\bottomrule	
	\end{tabular}
\caption{\small{Element-wise component analysis with ResNet backbone. GR: Global Reasoning, POM: Potential Object Mining, NSRM: Non-Salient Region Masking.}}
	\label{tab_element}	
\end{table}

\begin{figure*}[t]
	\begin{center}
		\includegraphics[width=\linewidth]{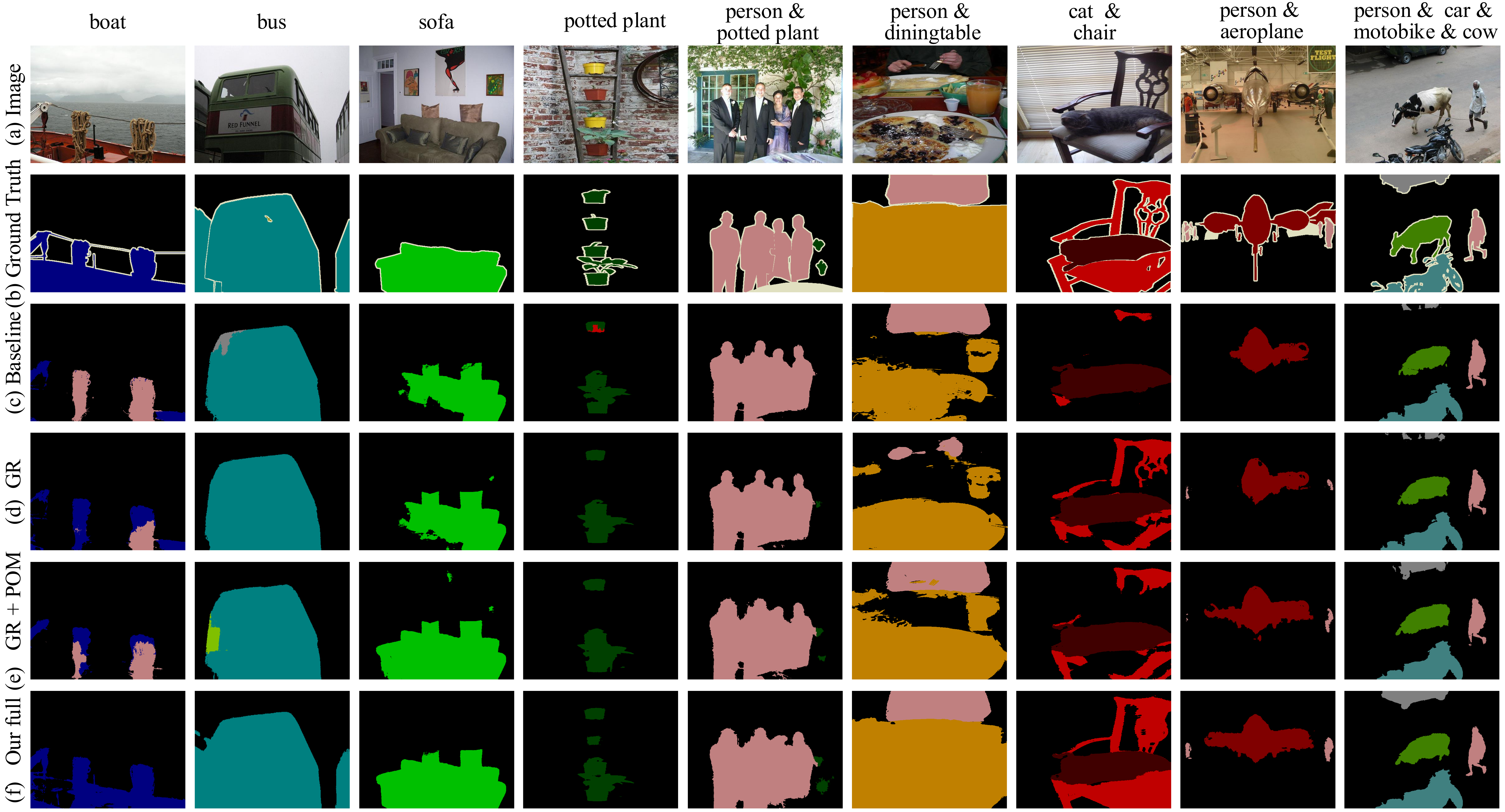}
	\end{center}
\vspace{-0.15cm}
	\caption{Example results on PASCAL VOC 2012 validation set. For each (a) image, we show the (b) ground truth, the result of (c) baseline, (d) GR, (e) GR + POM, and (f) our full method. The first four columns of images contain objects of one category, and the other five columns of images contain objects of two or more categories. Best viewed in color.}
	\vspace{-0.2cm}
	\label{fig_result}
\end{figure*}

\subsection{Ablation Studies}
\textbf{Element-Wise Component Analysis.}
In this part, we demonstrate the contribution of each component proposed in our approach for weakly supervised semantic segmentation. The experimental results on the validation set of Pascal VOC are given in Table \ref{tab_element}. We notice that, by leveraging the graph-based global reasoning unit (GR) to capture global relations among disjoint and distant regions, we can improve the segmentation result from 67.7\% to 68.8\%. By introducing our proposed potential object mining module (POM), we obtain another 0.2\% performance gain. Note that if we directly retrain the segmentation network with its prediction, the performance drops from 68.8\% to 68.7\%. In contrast, with our potential object mining module, retraining the segmentation network can further improve the result to 69.7\%. This highlights the importance of reducing the false-negative rate of pseudo labels. Our potential object mining module can directly improve the segmentation results and help to exert the self-correction ability of the segmentation network for a higher quality of pseudo labels. With our non-salient region masking module (NSRM), we further exploit the objects outside the conspicuous regions and improve the segmentation result to 70.4\%.

\begin{table}[t]
	
	\setlength{\tabcolsep}{4mm}
	\renewcommand\arraystretch{1.0}
	\centering
	\begin{tabular}{{l}*{1}{c}}
		\toprule
		Methods  & Val \\
		\midrule
		NSRM &70.4\\
		NSRM all &68.8\\
		NSRM - Object Expansion &70.2\\
		NSRM - Masking &70.0\\
		NSRM - dilation &68.5\\
		\bottomrule	
	\end{tabular}
\caption{\small{Ablation studies for NSRM with ResNet backbone.}}
	\label{tab_ab_nsrm}	
\end{table}

Some qualitative segmentation examples on the PASCAL VOC 2012 validation set can be viewed in Fig.~\ref{fig_result}. As can be seen, with the graph-based global reasoning unit (GR), the network can capture global relations and discover objects in disjoint and distant regions (\eg the potted plant in the fifth column and person in the eighth column). As shown in the last column, our method with the POM module can further discover the car outside the salient region. Besides, our full method with the NSRM module exerts the segmentation network's self-correction ability. It successfully predicts the bus and potted plants in the second and fourth columns, respectively. As shown in the fifth and eighth columns, our robust method further mines the objects outside the salient region.

\begin{figure*}[t]
	\begin{center}
		\includegraphics[width=\linewidth]{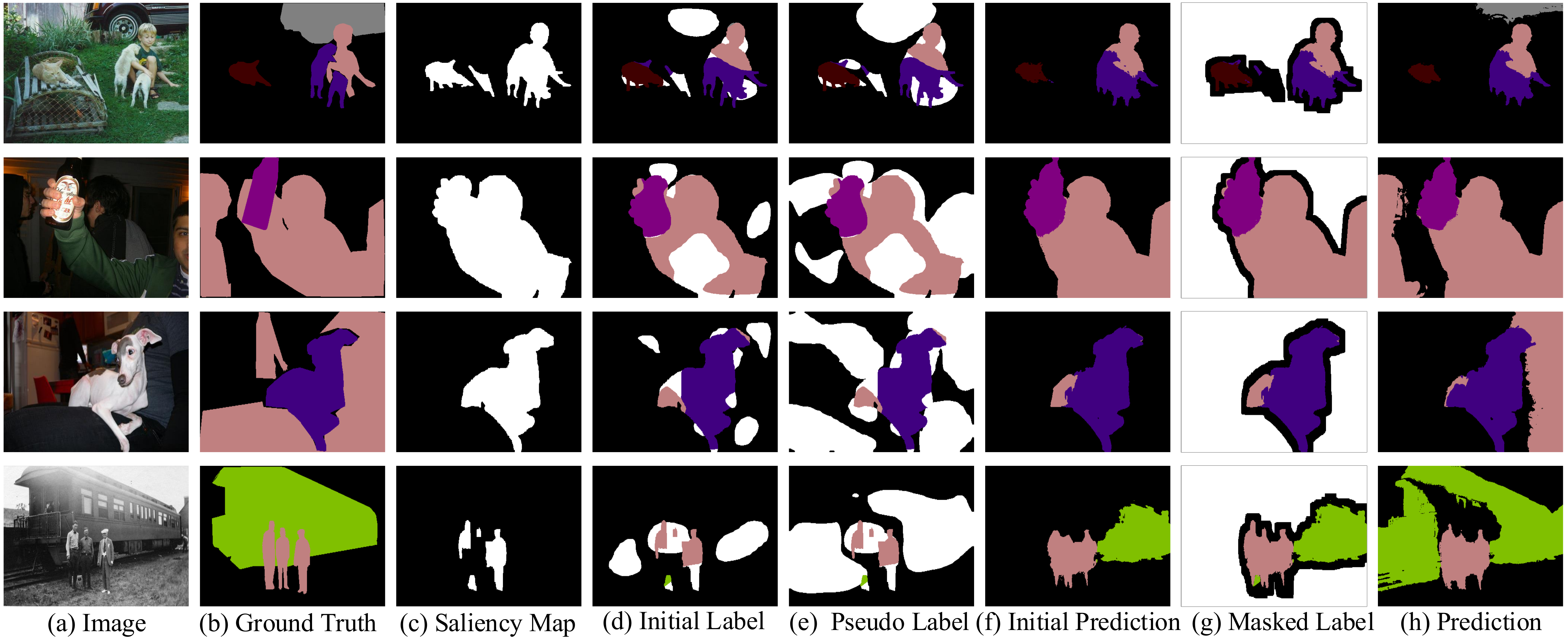}
	\end{center}
\vspace{-0.15cm}
	\caption{The evolution of labels for the PASCAL VOC 2012 training set. For each (a) image, we show the (b) ground truth, (c) saliency map, (d) initial label, (e) pseudo label, (f) initial prediction, (g) masked label, and (h) final prediction. Best viewed in color.}
	\label{fig_label}
	\vspace{-0.2cm}
\end{figure*}

We display the evolution of the labels we used for training the segmentation network in Fig.~\ref{fig_label}. As we can see, with our potential object mining module, we discover more object regions outside the salient area than the initial label. Our pseudo label can reduce the false gradients calculated for wrong annotations. By leveraging the initial prediction, our non-salient region masking module generates high-quality masked labels, allowing the segmentation model to mine the objects in the non-salient region further.

\textbf{Ablation Studies for NSRM.}
An in-depth study of our proposed NSRM module is presented in Table \ref{tab_ab_nsrm}. As we can see, if we apply NSRM to all images without our simple and complex image division, the results drop from 70.4\% to 68.8\%. This highlights the importance of treating simple and complex images differently. When masking out the non-salient region of pseudo labels for complex images during training, we need to rely on the rich background information provided by simple images. We notice that removing the object expansion operation will cause a 0.2\% performance drop. This shows that it is useful to utilize the pseudo labels to expand the object prediction within the salient region. Masking out the non-salient region of the pseudo label for training has a 0.4\% performance gain. This shows that the masking operation can encourage the segmentation network to exert its self-correction ability. Note that if we do not conduct the dilation operation for the object mask, the performance directly drops to 68.5\%. This highlights the importance of preserving the background area around the object. The background information, together with the object region, provides essential boundary knowledge for the network training.

\begin{figure}
	\centering
	\includegraphics[width=0.42\textwidth]{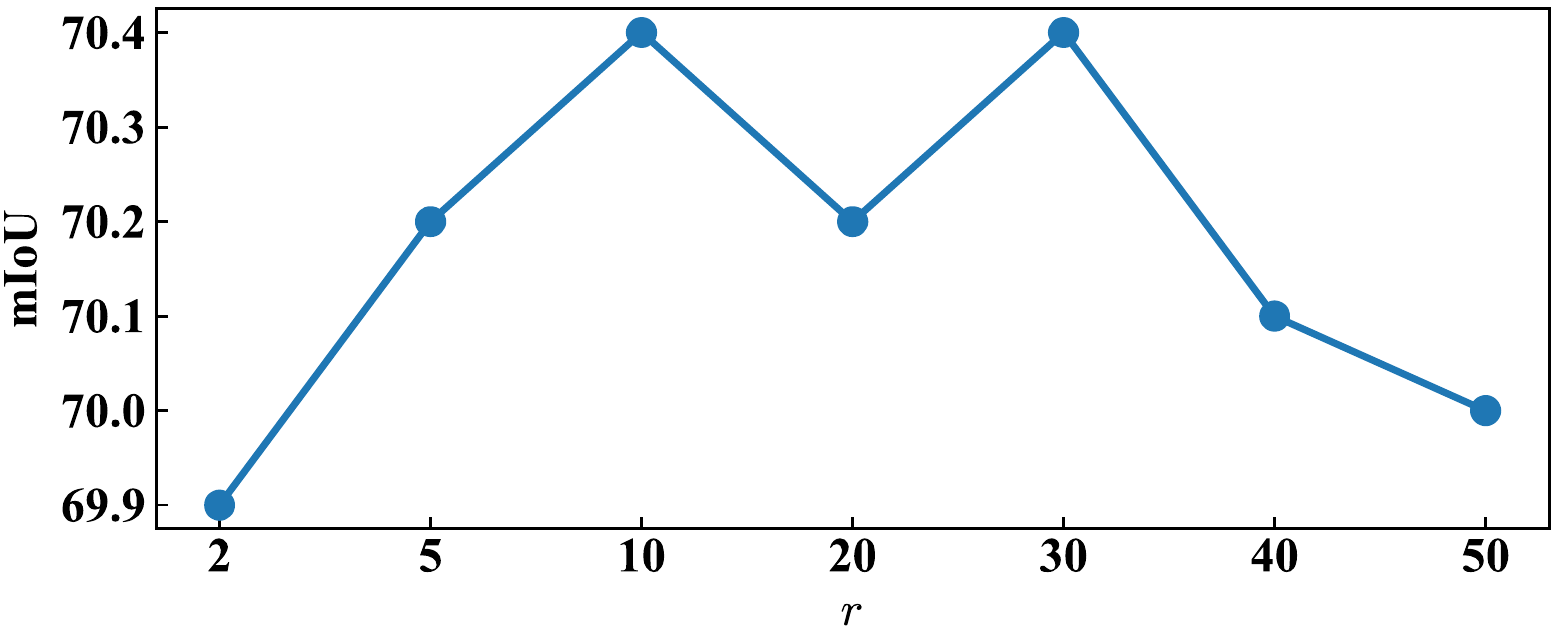}
	\vspace{-0.15cm}
	\caption{The parameter sensitivity of the dilation kernel size $r$ in the non-salient region masking module. Results are reported on the validation set using the ResNet backbone.}
	\vspace{-0.2cm}
	\label{fig_dilation}
\end{figure}

\begin{figure}
	\centering
	\includegraphics[width=0.42\textwidth]{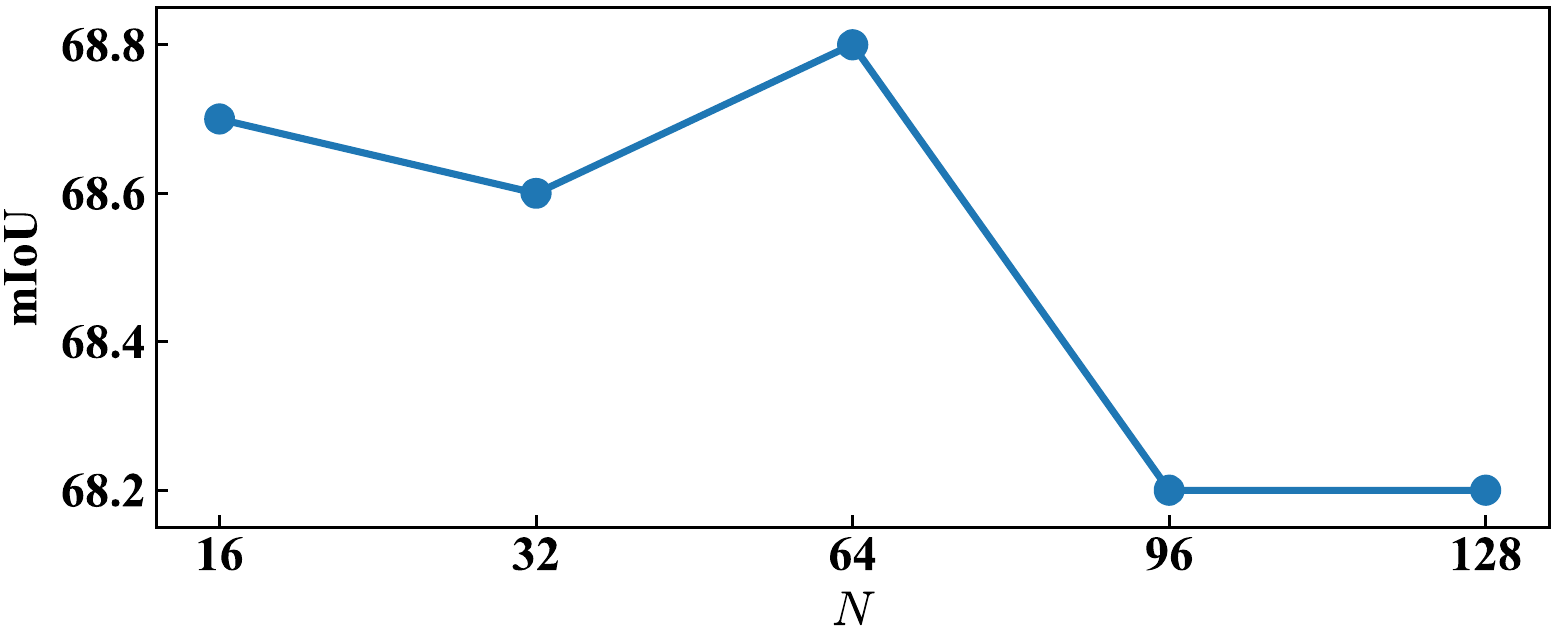}
	\vspace{-0.15cm}
	\caption{The parameter sensitivity of the number of feature nodes $N$ in the interaction space. Results are reported on the validation set using the ResNet backbone.}
	\vspace{-0.2cm}
	\label{fig_node}
\end{figure}

\textbf{Parameter Analysis.} 
For the dilation operation in the NSRM module, we conduct experiments to study the effect of the dilation kernel size $r$.  As shown in Fig.~\ref{fig_dilation}, we vary the kernel size $r$ over the range $\left \{2, 5, 10, 20, 30, 40, 50\right \}$. As we can see, we get better performance when the kernel size is between 5 and 30. A too large or small kernel size may not improve the results very much. We conjecture that a too large kernel size keeps too much background in the prediction, which hinders the object mining in the non-salient region. Meanwhile, a too small kernel size with little background blurs the boundary of objects, which impedes the training of the segmentation network. In our experiments, we empirically set $r$ = 30.

For the graph-based global reasoning unit, we conduct experiments to study the effect of the number of feature nodes $N$ in the interaction space. As shown in Fig.~\ref{fig_node}, we vary $N$ over the range $\left \{16, 32, 64, 96, 128\right \}$.  We notice that a too large number of the nodes $N$ may not improve the performance very much. In our experiments, we empirically set N = 64.

\section{Conclusions}
In this work, we proposed a non-salient region object mining approach for the task of weakly supervised semantic segmentation. Specifically, we introduced a graph-based global reasoning unit to help the classification network capture global relations among disjoint and distant regions. This can strengthen the network's ability to activate the objects scattered in the corners or near the edge of the image. To further mine objects outside the non-salient region, we proposed to exert the segmentation network's self-correction ability. A potential object mining module was proposed to reduce the false-negative rate in pseudo labels. Moreover, we proposed a non-salient region masking module for complex images to generate masked pseudo labels. Our non-salient region masking module helps further discover objects in the non-salient region. Extensive experiments on the PASCAL VOC 2012 dataset demonstrated the superiority of our proposed approach.

\section*{Acknowledgments}

This work was supported by the National Natural Science Foundation of China (No. 61976116) and Fundamental Research Funds for the Central Universities (No. 30920021135).

{\small
\bibliographystyle{ieee_fullname}
\bibliography{egbib}
}

\end{document}